\documentclass{article} 
\usepackage[final]{colm2025_conference}

\usepackage{microtype}
\usepackage{hyperref}
\usepackage{url}
\usepackage{booktabs}

\usepackage{lineno}

\usepackage{tcolorbox}
\usepackage{multicol}
\usepackage{caption}
\usepackage{amsmath}
\usepackage[noabbrev]{cleveref}
\usepackage{booktabs}
\usepackage{xcolor}
\usepackage{tabularx}
\usepackage{multirow}
\usepackage{siunitx}
\usepackage{url}
\usepackage{hyperref}

\definecolor{darkblue}{rgb}{0, 0, 0.5}
\hypersetup{colorlinks=true, citecolor=darkblue, linkcolor=darkblue, urlcolor=darkblue}

\title{Style over Substance: Distilled Language Models Reason \\ Via Stylistic Replication}

\author{Philip Lippmann\\
Delft University of Technology\\
\\
\And
Jie Yang\\
Delft University of Technology\\
}

\begin{document}

\ifcolmsubmission
\linenumbers
\fi

\maketitle

\begin{abstract}
%
Specialized reasoning language models (RLMs) have demonstrated that scaling test-time computation through detailed reasoning traces significantly enhances performance.
Although these traces effectively facilitate knowledge distillation into smaller, instruction-tuned models, the precise nature of transferred reasoning remains unclear.
In this study, we investigate to what extent distilled models internalize replicated stylistic patterns during reasoning. 
To this end, we systematically analyze reasoning traces, identifying structural and lexical patterns that characterize successful reasoning. 
We then introduce two new datasets -- a dataset of emergent reasoning traces and a synthetic dataset explicitly constructed to replicate these stylistic patterns -- to precisely examine their influence on distilled models' reasoning capabilities.
We find that models trained on the synthetic traces achieve comparable performance, indicating that distilled reasoning abilities rely significantly on surface-level patterns.
Surprisingly, we observe an increase in performance even when the synthetic traces are altered to lead to the wrong answer.
Our findings highlight how stylistic patterns can be leveraged to enhance LM reasoning across model families.
\end{abstract}


\section{Introduction}
\label{sec:intro}
Reasoning is fundamental to artificial intelligence, enabling systems to solve problems, make decisions, and explain outcomes. 
While traditional approaches to improving language model (LM) reasoning emphasize increased train-time compute~\citep{kaplan2020scalinglawsneurallanguage, hoffmann2022trainingcomputeoptimallargelanguage}, recent research highlights that scaling test-time compute through self-refinement is similarly effective~\citep{snell2024scalingllmtesttimecompute}. 
This insight has inspired specialized reasoning-focused LMs (RLMs), such as o1~\citep{o1} and R1~\citep{deepseekai2025deepseekr1incentivizingreasoningcapability}, which generate detailed reasoning traces of their thought process during inference.

\begin{figure}[t]
    \centering
    \includegraphics[width=\linewidth]{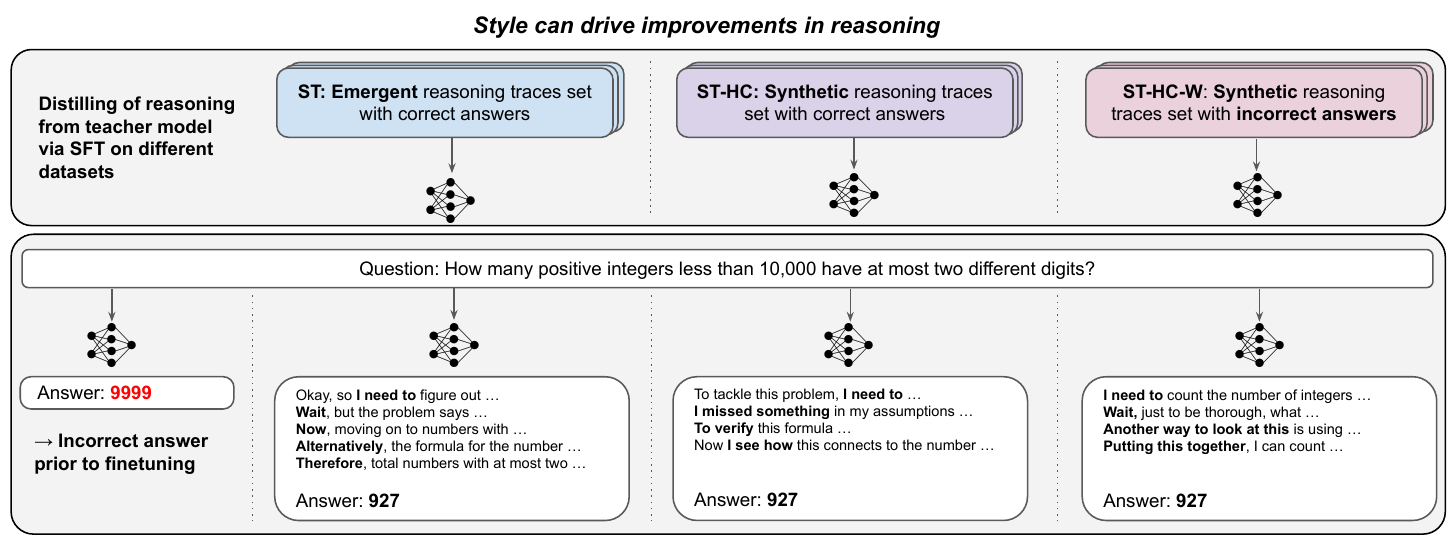}
    \caption{Reasoning trace style has noticeable influence on distilled model performance. We show how different types of reasoning traces -- emergent traces with correct answers, synthetic traces with correct answers, and synthetic traces with incorrect answers -- affect model performance after finetuning. All three approaches give the right answer due to improved reasoning capabilities compared to base models. Pivots highlighted in bold.}
    \label{fig:wrongright}
\end{figure}

Reasoning traces have proven effective for distillation~\citep{Schmidhuber, hinton2015distillingknowledgeneuralnetwork}, efficiently transferring sophisticated cognitive skills from RLMs to smaller, instruction-tuned models~\citep{sky_t1_2025}. 
Yet, the precise nature of the reasoning knowledge transferred remains poorly understood~\citep{zhu2023towards}. 
In particular, it is unclear whether distilled models genuinely internalize complex reasoning abilities or replicate superficial stylistic patterns from the original traces. 
This ambiguity echoes broader AI debates about whether language models genuinely understand content or simply engage in surface-level imitation~\citep{bender2020climbingnlunderstandingwithout, bender2021dangers, shanahan2022thinking, mirzadeh2025gsmsymbolic}. 
This uncertainty leads to a fundamental question: are we genuinely enhancing the reasoning capabilities of models, or merely teaching them to mimic domain-specific patterns that happen to improve performance on benchmarks?
Motivated by this ambiguity, we aim to establish whether \textbf{style is key to improvements in reasoning}, where style is characterized primarily by structural attributes such as trace length, lexical coherence, and backtracking frequency, rather than comprehension itself.

To examine this, we first systematically analyze successful reasoning traces produced by state-of-the-art RLMs, identifying recurrent structural and lexical patterns. 
This analysis, guided by cognitive science frameworks that characterize critical stages in human problem-solving~\citep{newell1972human}, reveals that effective reasoning traces consistently exhibit distinct \emph{metacognitive behaviors}. 
These behaviors are often signaled by lexical pivots -- markers such as ``Wait'' or ``What if'' -- that prompt reconsideration of assumptions or integration of new insights.

Based on these findings, we introduce two complementary datasets explicitly designed to clarify the role of style in improving reasoning. 
The first dataset, \textsc{SmolTraces} (\textsc{ST}), comprises verified question-answer pairs with sophisticated reasoning traces generated by a state-of-the-art RLM, displaying naturally emergent reasoning behaviors. 
The second dataset, \textsc{SmolTraces-HardCoded} (\textsc{ST-HC}), is synthetically constructed by embedding only the structural and lexical stylistic patterns identified earlier into reasoning traces generated by a standard LM without specialized reasoning capabilities. 
By comparing models trained on these datasets through supervised fine-tuning (SFT)~\citep{brown2020languagemodelsfewshotlearners}, we evaluate how stylistic consistency influences reasoning performance. 
Our experiments demonstrate that even stylistically consistent synthetic traces from a weaker model achieve comparable downstream reasoning performance, underscoring the importance of style as a critical factor in model training.

Further experiments contextualize these findings through targeted ablation studies, which examine the relative importance of trace correctness versus style. 
Remarkably, we find that stylistically consistent reasoning traces -- even those explicitly designed to lead to incorrect conclusions -- still substantially enhance downstream reasoning performance over the base model, as shown in \cref{fig:wrongright}. 
These results underscore that stylistic consistency significantly influences LM reasoning capabilities, providing an explanation for the effectiveness of reasoning distillation from RLMs to regular LMs.

In summary, our work makes two major contributions: (1) we demonstrate that distilled reasoning improvements rely heavily on stylistic patterns present in reasoning traces; and (2) we identify specific structural and lexical features indicative of successful reasoning. 
Collectively, these contributions deepen our understanding of how language models perform reasoning tasks, and the datasets we release can serve as valuable resources for future research into synthetic data generation and fine-tuning methodologies that explicitly target the relationship between reasoning trace style and substance.


\begin{figure*}[t]
    \centering
    \includegraphics[width=\textwidth]{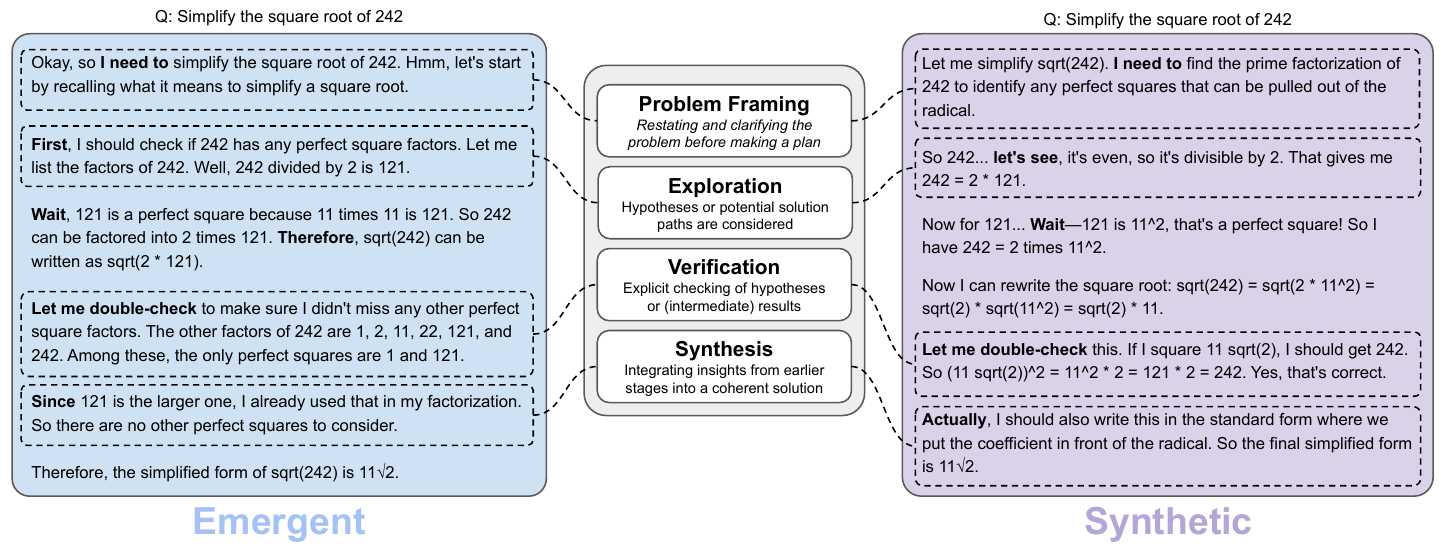}
    \caption{Comparison of emergent and synthetic reasoning traces for solving the same problem. The left side shows a reasoning trace generated by an RLM, while the right side displays a synthetic trace created using our hard-coded template with predefined pivots. Both approaches follow similar cognitive stages (center): problem framing, exploration, verification, and synthesis. The dashed boxes highlight examples of each stage, demonstrating that synthetic traces can effectively replicate the style of emergent reasoning.}
    \label{fig:workflow}
\end{figure*}


\section{Background}
\label{sec:background}

\noindent\textbf{Chain-of-Thought}
Early approaches to elicit reasoning from LMs, such as Chain-of-Thought (CoT)~\citep{wei2023chainofthoughtpromptingelicitsreasoning}, demonstrate that intermediate reasoning steps are key for LMs to improve their problem-solving abilities~\citep{zhang2022automaticchainthoughtprompting}.
While CoT improves performance on reasoning tasks, it primarily focuses on generating a linear sequence of steps towards a solution.
On the other hand, RLM reasoning traces (sometimes called ``long CoT'') -- which are the focus of this paper -- differ from CoT as they do not just try to build toward the solution linearly, but instead actively backtrack, verify, and explore different lines of thinking.

\noindent\textbf{Reasoning traces}
Reasoning traces are semi-structured textual representations that capture a model's thought process while working toward the solution of a problem during inference.
Here, additional tokens are generated before the final answer is given to reason about the problem.
These traces typically include explicit markers of metacognition, such as planning statements, hypothesis testing, and self-correction.
For an example of a full reasoning trace, see \cref{sec:app_ex}.
A key characteristic of effective reasoning traces is the presence of \emph{pivots}, points where the model explicitly moves between different categories of metacognition.
We delve into the specific types of pivots in \cref{sec:structure_pivots}.

\noindent\textbf{Finetuning on reasoning traces}
Recent research has demonstrated that finetuning language models on reasoning traces significantly enhances their reasoning capabilities~\citep{huang2024o1replicationjourney, sky_t1_2025, deepseekai2025deepseekr1incentivizingreasoningcapability}.
In this approach, detailed reasoning traces generated by RLMs are used as training data to transfer structured reasoning behaviors into smaller, instruction-tuned models via SFT~\citep{ min2024imitateexploreselfimprovereproduction}.
Models finetuned on these traces consistently outperform those trained only on final answers or simpler step-by-step solutions, suggesting that the explicit structure and content of reasoning traces play a critical role in improving model performance~\citep{xu2025redstardoesscalinglongcot, bespoke_stratos}.
Despite these successes, the precise mechanisms underlying the effectiveness of reasoning trace distillation remain unclear, particularly regarding the balance between structural stylistic cues and the cognitive complexity of the reasoning itself.


\section{Hard-coding reasoning traces to approximate emergent ones}
\label{sec:method}

In examining whether distilled models internalize genuine reasoning capabilities or primarily benefit from structural and lexical patterns, we investigate the relationship between \textit{style} and \textit{substance} in reasoning traces.
Here, style encompasses the structural and lexical features while substance refers to factual correctness and semantic content.
To examine stylistic influence, we develop a methodology that replicates the structural patterns of successful reasoning while varying content, allowing us to assess style's contribution to reasoning distillation.
We hypothesize that emergent reasoning behaviors can be effectively approximated by encoding the metacognitive pivots characteristic of RLM traces into synthetic reasoning traces.
We define \textit{emergent traces} as those naturally produced by RLMs after training via reinforcement learning, while \textit{synthetic traces} refer to our approximations that incorporate the stylistic elements we identify as key, but originate from standard LMs without specialized reasoning capabilities.
This approach reduces reliance on costly RLM inference\footnote{At time of writing, the difference in API costs per token between flagship LMs and RLMs approach an order of magnitude from the same provider: input/output pricing of \$2.50/\$10.00 for GPT-4o compared to \$15.00/\$60.00 for o1 \url{https://openai.com/api/pricing/} [Accessed: 2025-02-21]} while enabling us to control stylistic components that potentially drive performance improvements in distilled models.

\subsection{Reasoning trace structure and pivot types} 
\label{sec:structure_pivots}
\noindent\textbf{Trace analysis}
Effective reasoning traces exhibit systematic structural patterns reflective of human-like problem-solving strategies. 
Cognitive science literature characterizes human reasoning as a structured process comprising distinct stages: \emph{problem framing}, \emph{hypothesis exploration}, \emph{verification}, and \emph{synthesis}~\citep{newell1972human}. 
Guided by this framework, we systematically analyze 17K successful reasoning traces produced by a state-of-the-art RLM (see \cref{sec:app_tracean} for details) and find that effective RLM-generated reasoning traces consistently align with these cognitive stages.
Specifically, reasoning traces begin with explicit \emph{problem framing}, restating and clarifying key aspects of the problem leading to a plan, followed by an \emph{exploration} stage in which hypotheses or potential solution paths are considered. 
This is complemented by a \emph{verification} stage, characterized by explicit checking of hypotheses or intermediate results, culminating in a final \emph{synthesis}, integrating insights from earlier stages into a coherent solution.

\noindent\textbf{Pivot types}
Crucially, effective reasoning traces are rarely linear; instead, they frequently revisit previous stages to correct errors, validate assumptions, or explore alternative strategies. 
This non-linear metacognitive behavior is operationalized through \emph{pivots} -- lexical markers signaling explicit shifts between reasoning stages.
Our analysis identifies four primary pivot categories, each aligning closely with a corresponding reasoning stage: 
(1) \emph{Realization pivots}, such as ``Wait'' or ``Oh,'' signal recognition of errors or oversights during the exploration stage; (2) \emph{Verification pivots}, initiated by phrases such as ``Let me check,'' explicitly validate intermediate hypotheses; (3) \emph{Exploration pivots}, introduced with phrases such as ``What if'' or ``Another approach,'' prompt the consideration of alternative solution paths; and (4) \emph{Integration pivots}, typically signaled by expressions such as ``Now I see how,'' synthesize previously explored ideas into a coherent final solution.
We visualize what these stages look like in practice for emergent and synthetic traces in \cref{fig:workflow}

\begin{figure*}[t]
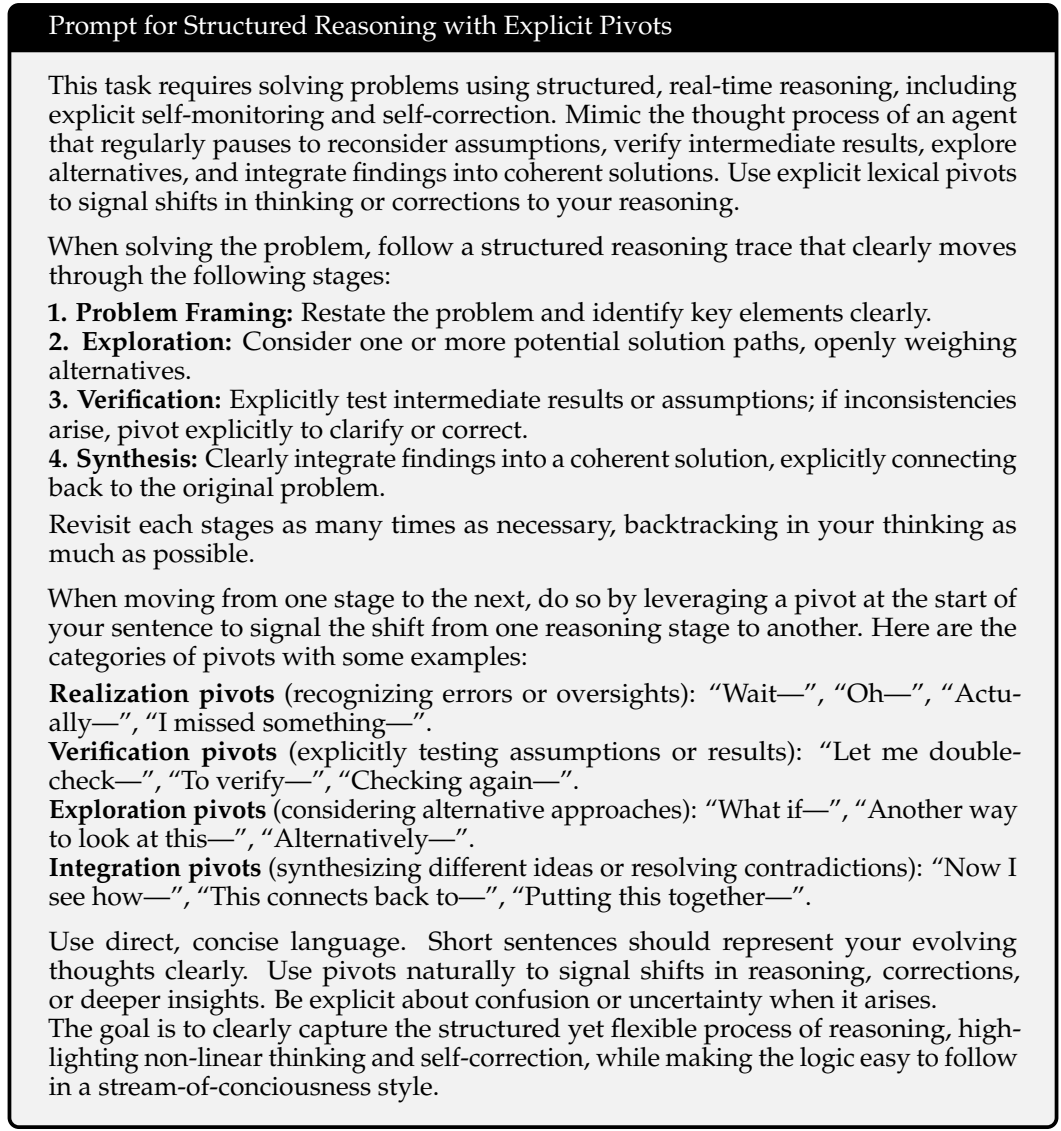

\begin{tcolorbox}[colback=gray!10, colframe=black, title=Prompt for Structured Reasoning with Explicit Pivots, width=\textwidth]
This task requires solving problems using structured, real-time reasoning, including explicit self-monitoring and self-correction. Mimic the thought process of an agent that regularly pauses to reconsider assumptions, verify intermediate results, explore alternatives, and integrate findings into coherent solutions. Use explicit lexical pivots to signal shifts in thinking or corrections to your reasoning.

\medskip

When solving the problem, follow a structured reasoning trace that clearly moves through the following stages:

\smallskip
\textbf{1. Problem Framing:} Restate the problem and identify key elements clearly.

\textbf{2. Exploration:} Consider one or more potential solution paths, openly weighing alternatives.

\textbf{3. Verification:} Explicitly test intermediate results or assumptions; if inconsistencies arise, pivot explicitly to clarify or correct.

\textbf{4. Synthesis:} Clearly integrate findings into a coherent solution, explicitly connecting back to the original problem.

\smallskip

Revisit each stages as many times as necessary, backtracking in your thinking as much as possible.

\medskip

When moving from one stage to the next, do so by leveraging a pivot at the start of your sentence to signal the shift from one reasoning stage to another. Here are the categories of pivots with some examples:

\smallskip

\textbf{Realization pivots} (recognizing errors or oversights): ``Wait—'', ``Oh—'', ``Actually—'', ``I missed something—''.  

\textbf{Verification pivots} (explicitly testing assumptions or results): ``Let me double-check—'', ``To verify—'', ``Checking again—''.  

\textbf{Exploration pivots} (considering alternative approaches): ``What if—'', ``Another way to look at this—'', ``Alternatively—''.  

\textbf{Integration pivots} (synthesizing different ideas or resolving contradictions): ``Now I see how—'', ``This connects back to—'', ``Putting this together—''.

\medskip

Use direct, concise language. Short sentences should represent your evolving thoughts clearly. Use pivots naturally to signal shifts in reasoning, corrections, or deeper insights. Be explicit about confusion or uncertainty when it arises.  

The goal is to clearly capture the structured yet flexible process of reasoning, highlighting non-linear thinking and self-correction, while making the logic easy to follow in a stream-of-conciousness style.
\end{tcolorbox}
\caption{\small The prompt used to guide GPT-4o for generating the synthetic \textsc{ST-HC} dataset. This prompt explicitly defines the four key pivot types (Realization, Verification, Exploration, Integration) and mandates adherence to the four reasoning stages (Problem Framing, Exploration, Verification, Synthesis) derived from our analysis of emergent RLM traces (\cref{sec:structure_pivots}). Its goal is to enforce specific stylistic patterns, including non-linear thinking and explicit self-correction, characteristic of effective reasoning.}
\label{fig:prompt}
\end{figure*}

Our analysis reveals that successful reasoning traces commonly employ multiple pivot types (96.1\% contain at least three pivot categories, additional information per type given in \cref{sec:app_tracean}), whereas unsuccessful traces frequently lack such pivots or exhibit limited diversity.
This underscores the critical importance of structured, metacognitive transitions in effective reasoning.
Motivated by these insights, we explicitly encode these structural and lexical patterns into a reasoning template to facilitate the generation of synthetic reasoning traces that capture RLM-like reasoning behaviors.
We show the associated prompt structure in \cref{fig:prompt}, which we subsequently use to guide reasoning trace generation (\cref{sec:datagen}), enabling us to systematically control and isolate stylistic reasoning elements in order to examine their impact on reasoning performance.
First, the pivot categories are explicitly defined, followed by the general stages of problem solving that they correspond to.
To generate the synthetic data, in addition to this prompt, the teacher LM is given the question and instructions on how to format its answer.

\subsection{Reasoning trace data generation}
\label{sec:datagen}

\noindent\textbf{Collect seed data}
Initially we curate seed data consisting of questions and their correct answers, ensuring that the accuracy of the eventual synthetic data can be verified.
While previous works on RLM distillation often focus exclusively on math~\citep{huang2024o1replicationjourney}, we aim to cover a wider range of additional domains that benefit from reasoning, such as coding, science, and logic.
Specifically, for questions from several scientific domains, we select OlympicArena~\citep{huang2024olympicarenabenchmarkingmultidisciplinecognitive}.
For logic and coding, we select AGIEval~\citep{zhong2023agievalhumancentricbenchmarkevaluating} and LiveCodeBench v4~\citep{jain2024livecodebench}, respectively.
We select NuminaMATH~\citep{numina_math_datasets}, where we randomly select a subset of 20,000 samples, and OmniMath~\citep{gao2024omnimathuniversalolympiadlevel} as sources of quantitative reasoning problems for our seed data, resulting a total of 31,586 question-answer pairs.

\noindent\textbf{Generate synthetic traces} 
We use the seed data to generate high-quality synthetic reasoning traces via state-of-the-art RLMs and LMs.
For \textsc{ST}, we choose R1 as, at the time of writing, it is the best performing RLM that provides its full reasoning traces as part of its response.
For \textsc{ST-HC}, we choose GPT-4o as our teacher model -- using the prompt structure specified in \cref{fig:prompt}.
We perform up to five rollouts per seed sample, discarding incorrect responses, stopping if the model provides the correct answer.
The rollouts are done in a zero-shot manner, i.e. we provide only the question to the model and do not keep previous attempts as context.

\noindent\textbf{Filtering synthetic samples}
First, we filter out samples that are of short length (less than 50 tokens for the entire trace) to prioritize sample quality, as shorter samples typically contain few pivots and are therefore not as impactful in training.
In an effort to provide a fairer comparison, we align the datasets to the same number of samples by downsizing the larger dataset.
As the final \textsc{ST-HC} dataset contained fewer samples (N=18,242), we downsample the larger \textsc{ST} dataset to match this size, randomly removing questions not present in \textsc{ST-HC} and ensuring both datasets used for finetuning contain an equal number of samples.
This results in a final 18K samples for both \textsc{ST} and \textsc{ST-HC}, each in the form of a triple (question, reasoning trace, answer).
We provide a more detailed overview of the resulting datasets in \cref{sec:appendix}.

\subsection{Experimental details}
\label{sec:training}

\noindent\textbf{Model finetuning}
We finetune a range of already instruction-tuned base LMs using our contributed reasoning datasets.
For this, we select recent models of different families and sizes, namely: Llama 3.2 3B~\citep{metallama}, Ministral 8B~\citep{jiang2023mistral7b}, and Qwen2.5 32B~\citep{qwen2025qwen25technicalreport}.
We choose these models as they are all high performing for their parameter count, come with open-source weights, and have permissive licenses.
The learning rate used during SFT varies for each LM in line with the model's parameter count.
For 3B models we use a peak learning rate of $6 \times 10^{-5}$, for 8B models we use $4 \times 10^{-5}$, and for 32B models we use $1 \times 10^{-5}$.
For all models we use an effective batch size of 16.
All models are trained for five epochs using a linear warmup for the first 10\% of steps followed by cosine annealing.
We use the AdamW optimizer~\citep{loshchilov2019decoupledweightdecayregularization} with $\beta_1 = 0.9$ and $\beta_2 =0.95$ and a weight decay of $1 \times 10^{-4}$.
Training is performed on a system comprising 8 Nvidia H100 GPUs using \texttt{bfloat16} precision.

\noindent\textbf{Baselines}
To contextualize our findings, we evaluate (1) the base instruction-tuned model, (2) the model after SFT on \textsc{ST}, and (4) the model after SFT on \textsc{ST-HC}.
Additionally, to gauge the impact of the particular style replication we propose versus distilling from regular CoT, we add another baseline, where we instruct the generating LLM to think step-by-step (\textsc{SBS}), following~\citet{kojima2023largelanguagemodelszeroshot}. 
We then use the resulting CoT for distillation -- similarly to how we use the emergent (\textsc{ST}) and synthetic (\textsc{ST-HC}) reasoning traces.
This comprehensive evaluation allows us to isolate the impact of our hard-coded reasoning approach across different model sizes and compare it against the reasoning capabilities of the model itself.
For completeness, we evaluate the teacher models used to generate our datasets as well.

\noindent\textbf{Evaluation}
To evaluate the reasoning capabilities of all models, we select challenging, widely used benchmarks that test reasoning capabilities.
These include MATH500~\citep{lightman2023letsverifystepstep}, AIME2024, and GPQA~\citep{rein2024gpqa}; covering math and a range of scientific domains.
Specifically, MATH500 and AIME2024 feature 500 and 30 competition math problems, respectively.
GPQA consists of 198 questions from a range of scientific fields such as Biology and Chemistry. 
We focus only on the hardest (``Diamond'') subset of this particular dataset.
%
%


\section{Results and discussion}
\label{sec:experiments}

\subsection{Hard-coded reasoning results}
\label{sec:results}

The performance across all combinations of models and datasets evaluated in our study is presented in \cref{tab:performance}.
Our results clearly indicate that models finetuned with structured reasoning traces, whether emergent (\textsc{ST}) or synthetic (\textsc{ST-HC}), consistently and significantly outperform their baseline instruction-tuned counterparts across all benchmarks.
Notably, even the smaller 3B and 8B parameter models exhibit substantial performance improvements. 
For instance, the Llama 3.2 3B model gains over 31 absolute percentage points on MATH500 when finetuned on \textsc{ST}, and the Ministral 8B model sees its AIME2024 score more than triple with either \textsc{ST} or \textsc{ST-HC} finetuning. 
These findings demonstrate that smaller models can markedly benefit from structured reasoning finetuning, challenging previous suggestions of minimal improvements for models of this scale~\citep{sky_t1_2025}.
Comparing the two fine-tuning approaches, we observe that models trained on synthetic traces achieve performance that approaches, and in some cases matches those trained on emergent RLM traces. 
%
This demonstrates that replicating the style of reasoning, even using a less capable teacher model guided by our prompt, is highly effective for distilling reasoning capabilities. 
%
%
Finally, the results show that while generating synthetic data with a \textsc{SBS} prompt is beneficial compared to the base model, there is a significant performance gap between \textsc{SBS} and our \textsc{ST} and \textsc{ST-HC} methods.
This demonstrates that the performance improvements are indeed substantially driven by the specific stylistic patterns we identified, rather than solely by distilling the generating LLM's general reasoning abilities.

\begin{figure}[t]
    \centering
    \includegraphics[width=\linewidth]{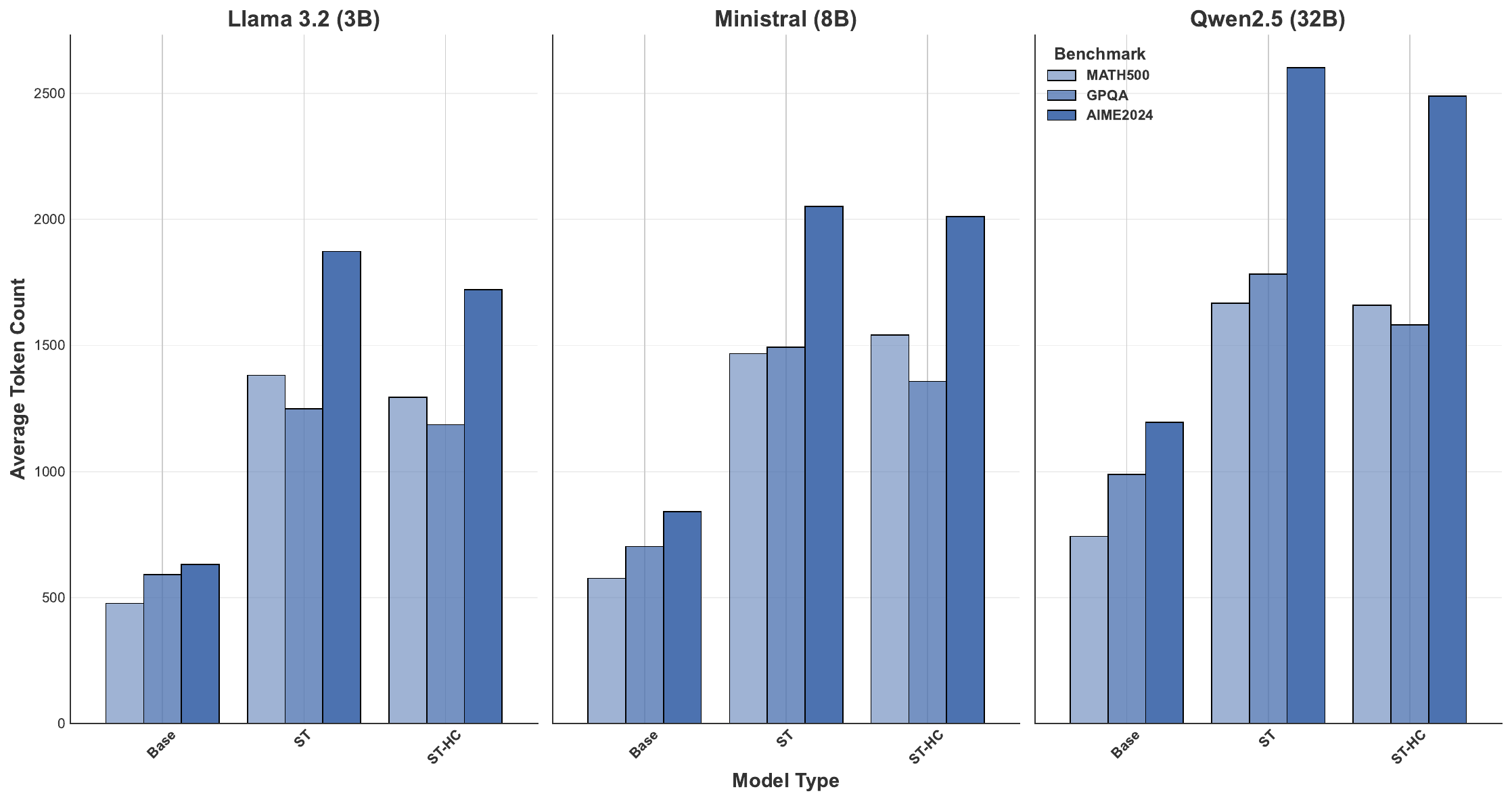}
    \caption{Average token count by model family and training dataset across evaluations.}
    \label{fig:tokens}
\end{figure}

When analyzing the thinking process during evaluation, we observe a clear correlation between successful reasoning and longer reasoning traces, as illustrated in \cref{fig:tokens}.
Models finetuned on either emergent or synthetic reasoning traces consistently produce substantially longer outputs compared to their respective base models across all evaluation benchmarks.
This suggests that adopting and replicating a structured, elaborate reasoning style -- whether learned from emergent traces or via synthetic ones -- is a key mechanism driving the enhanced downstream reasoning capabilities we observed.
Notably, \cref{fig:tokens} shows that finetuning on \textsc{ST} yields slightly longer reasoning traces than \textsc{ST-HC}, which correlates with their relative performance to one another -- corroborating that a larger number of tokens spent thinking typically correlates with improved reasoning~\citep{muennighoff2025s1simpletesttimescaling}.


\begin{table*}[t]
\centering
\small
\begin{tabular*}{\textwidth}{@{\extracolsep{\fill}} l l c c c c @{}}
\toprule
\textbf{Model} & \textbf{Variant} & \textbf{Params} & \textbf{MATH500} & \textbf{AIME2024} & \textbf{GPQA (D)} \\
\midrule
\multirow{3}{*}{\textbf{Llama 3.2}}
& Base            & 3B  & 36.4 & 6.7  & 26.3   \\
& \textsc{SBS}    & 3B  & 45.8 & 10.0 & 28.3   \\
& \textsc{ST}     & 3B  & 68.4 & 23.3 & 31.3  \\
& \textsc{ST-HC}  & 3B  & 64.2 & 16.7 & 29.3  \\
\midrule
\multirow{3}{*}{\textbf{Ministral}}
& Base            & 8B  & 52.8 & 10.0  & 28.8  \\
& \textsc{SBS}    & 8B  & 60.6 & 16.7 & 31.3   \\
& \textsc{ST}     & 8B  & 78.2 & 33.3 & 38.9  \\
& \textsc{ST-HC}  & 8B  & 77.0 & 33.3 & 34.8  \\
\midrule
\multirow{3}{*}{\textbf{Qwen2.5}}      
& Base            & 32B & 76.8 & 16.7 & 49.0  \\
& \textsc{SBS}    & 32B & 78.2 & 20.0 & 49.5   \\
& \textsc{ST}     & 32B & 89.0 & 53.3 & 56.1  \\
& \textsc{ST-HC}  & 32B & 83.4 & 46.7 & 53.0  \\
\midrule\midrule
\multirow{3}{*}{\textbf{Teacher Models}}   
& R1                    & 671B & 96.8 & 76.7 & 71.7  \\
& GPT-4o                & -    & 75.4 & 13.3 & 53.0  \\
& GPT-4o \textsc{HC}    & -    & 81.2 & 16.7 & 55.1  \\
\bottomrule
\end{tabular*}
\caption{Performance comparison of language models finetuned on reasoning traces. We compare base models against versions finetuned on \textsc{SBS} (step-by-step thinking GPT-4o), \textsc{ST} (emergent traces from R1), and \textsc{ST-HC} (synthetic traces from GPT-4o with the prompt in \cref{fig:prompt}). This evaluates the impact of reasoning trace style on downstream performance across model scales and benchmarks. GPT-4o \textsc{HC} refers to the base LM prompted with our structure from \cref{fig:prompt}. All results are pass@1 accuracy as a percentage.}
\label{tab:performance}
\end{table*}



\subsection{Ablative study}
\label{sec:abl}

To further isolate the impact of reasoning style versus factual correctness, we conduct an ablation study using two modified datasets. 
First, we create \textsc{ST-HC-W} by adapting the synthetic \textsc{ST-HC} traces to retain their stylistic structure while leading to incorrect final answers generated by GPT-4o-mini (details given in \cref{sec:app_abla}). 
Second, we create \textsc{ST-NT}, which contains only the question-answer pairs from the original \textsc{ST} data, removing the reasoning traces entirely.
The results, presented in \cref{tab:performance_abla}, reveal several key insights. 
Models finetuned on \textsc{ST-HC-W} consistently outperform the base instruction-tuned models across all evaluation benchmarks. 
This demonstrates that learning the stylistic patterns of reasoning enhances problem-solving capabilities even when the training data's final conclusion is incorrect.
However, \textsc{ST-HC-W} models perform noticeably worse than those trained on the stylistically similar but factually correct \textsc{ST-HC} dataset, confirming the value of accurate data.
Unsurprisingly, models trained on \textsc{ST-NT} show only modest gains over the base models, as they do not learn to generate additional thinking tokens in the style of an RLM during inference.

These ablations confirm that while factual correctness is important for optimal performance, the stylistic patterns inherent in reasoning traces play a critical role in enhancing the reasoning abilities of distilled models. 
Crucially, we do not claim that style alone improves reasoning as the body of the traces still contains correct reasoning up until the answer, but rather emphasize that style is essential for improving reasoning capabilities.
Thus, unlike traditional distillation methods relying predominantly on sample correctness, our results suggest that explicitly transferring a specific output structure is important.


\section{Related work}
\label{sec:related}

\noindent\textbf{Language model reasoning}
Language model reasoning has received increasing interest in recent years~\citep{cobbe2021trainingverifierssolvemath}.
Initially, LMs were conditioned on reasoning examples during pretraining, post-training or in-context to improve their reasoning capabilities~\citep{zhang2022automaticchainthoughtprompting,wang2023selfconsistencyimproveschainthought, wei2023chainofthoughtpromptingelicitsreasoning, li2024common7blanguagemodels}, but test-time scaling~\citep{snell2024scalingllmtesttimecompute, muennighoff2025s1simpletesttimescaling} has introduced a new paradigm for improving LM reasoning.
Further, \citet{deepseekai2025deepseekr1incentivizingreasoningcapability} explore training RLMs with reinforcement learning, finding that they mimic human reasoning processes like self-reflection and verification.
The resulting RLMs provide reasoning traces as training data to enhance LM reasoning ability.
Our work demonstrates that the structural and lexical patterns in these traces play a significant role in improving reasoning performance, offering insights into what is actually transferred during reasoning distillation.

\noindent\textbf{Generalization in language models}
Generalization in language models implies the ability to tackle unseen problems rather than simply reciting training data~\citep{kang2024learningdynamicsrevealgeneralization}.
LMs struggle with problems that differ from their training distribution, often closely following observed patterns down to individual terms~\citep{razeghi2022impactpretrainingtermfrequencies}.
Recent research shows LMs rely heavily on memorized patterns rather than developing generalizable reasoning capabilities~\citep{schwarzschild2024rethinking}.
This pattern-matching behavior is especially evident in mathematical reasoning, where LMs show variance across different instantiations of the same question and declining performance when only numerical values change~\citep{mirzadeh2025gsmsymbolic}.
Data contamination has also been identified as a source of apparent but false generalization~\citep{jiang2024investigatingdatacontaminationpretraining}.
Our work extends this discussion by investigating how stylistic patterns in reasoning traces influence model performance, revealing that structural elements of reasoning may be as important as factual content for enhancing problem-solving capabilities.

\noindent\textbf{Reasoning distillation}
Distillation has long been used to improve various aspects of machine learning models~\citep{Schmidhuber, hinton2015distillingknowledgeneuralnetwork, sanh2020distilbertdistilledversionbert}.
For reasoning tasks with verifiable solutions, researchers have implemented rejection sampling methodologies that extract and validate advanced models' reasoning processes~\citep{zelikman2022star}.
More recently, significant performance improvements have been achieved through SFT on synthetic datasets generated by superior LMs~\citep{gunasekar2023textbooksneed}.
With the emergence of RLMs, reasoning performance of instruction-tuned LMs can be substantially enhanced by finetuning on high-quality reasoning traces~\citep{deepseekai2025deepseekr1incentivizingreasoningcapability}.
Multiple efforts have demonstrated success in distilling RLMs via SFT on reasoning traces that contain step-by-step thinking~\citep{min2024imitateexploreselfimprovereproduction, huang2024o1replicationjourney, sky_t1_2025, bespoke_stratos, xu2025redstardoesscalinglongcot}.
Our work attempts to decompose what is actually being transferred during this distillation process, revealing that stylistic elements of reasoning traces contribute significantly to performance gains independent of their factual correctness.


\begin{table*}[t]
\centering
\small
\begin{tabular*}{\textwidth}{@{\extracolsep{\fill}} l l c c c c @{}}
\toprule
\textbf{Model} & \textbf{Variant} & \textbf{Parameters} & \textbf{MATH500} & \textbf{AIME2024} & \textbf{GPQA Diamond} \\
\midrule
\multirow{2}{*}{\textbf{Llama 3.2}}  
& \textsc{ST-HC-W}              & 3B  & 48.2 & 10.0  & 28.4   \\
& \textsc{ST-NT}                & 3B  & 40.6 &  6.7  & 26.9   \\
\midrule
\multirow{2}{*}{\textbf{Ministral}}
& \textsc{ST-HC-W}              & 8B  & 62.8 & 20.0  & 29.9   \\
& \textsc{ST-NT}                & 8B  & 56.2 & 13.3  & 30.0   \\
\midrule
\multirow{2}{*}{\textbf{Qwen2.5}}      
& \textsc{ST-HC-W}              & 32B & 80.2 & 26.7  & 51.3   \\ 
& \textsc{ST-NT}                & 32B & 78.8 & 20.0  & 49.8   \\ 
\bottomrule
\end{tabular*}
\caption{Ablation study results evaluating the distinct contributions of reasoning trace style and answer correctness. We compare model performance after finetuning on: (1) \textsc{ST-HC-W}, featuring stylistically consistent synthetic traces from \textsc{ST-HC} but deliberately leading to \textit{incorrect} answers, and (2) \textsc{ST-NT}, containing only the question-answer pairs from \textsc{ST} \textit{without} reasoning traces. All results are pass@1 accuracy as a percentage.}
\label{tab:performance_abla} 
\end{table*}

\section{Conclusion}
\label{sec:conclusion}
Our work challenges existing assumptions by demonstrating that improvements in reasoning capabilities in distilled language models can be driven by stylistic replication.
By analyzing reasoning traces through the lens of established principles from cognitive science on how humans solve problems, we identified structural patterns and key lexical markers (\emph{pivots}) characteristic of effective LM reasoning. 
We validated these findings by synthetically replicating the stylistic patterns found in emergent reasoning traces using a weaker model, achieving comparable downstream reasoning performance after distillation.
Ablation studies confirmed that even reasoning traces with incorrect solutions enhance downstream performance, highlighting the effectiveness of stylistic mimicry in distilled models. 
Collectively, our results contribute to a deeper understanding of reasoning in language models and offer practical insights into efficient finetuning strategies and data generation practices.

\bibliography{colm2025_conference}
\bibliographystyle{colm2025_conference}

\appendix

\newpage
\section{Emergent Reasoning Trace Example}
\label{sec:app_ex}
We give an example of a reasoning trace in \cref{fig:trace-example}.
Here, a RLM answers a mathematical puzzle question, initially producing an extensive reasoning trace with frequent backtracking, followed by giving the final correct answer.

\begin{figure*}[h]
\begin{tcolorbox}[colback=gray!10, colframe=black, title=Reasoning Trace Example, sharp corners, width=\textwidth]
\textbf{Question:} A juggler can juggle 16 balls. Half of the balls are golf balls, and half of the golf balls are blue. How many blue golf balls are there?

\noindent\rule{\textwidth}{0.5pt} 

\textbf{Reasoning:} Okay, let's see. The problem says a juggler can juggle 16 balls. Half of them are golf balls, and half of the golf balls are blue. I need to find out how many blue golf balls there are. Alright, let me break this down step by step.

First, the total number of balls is 16. Half of those are golf balls. So, half of 16 would be... 16 divided by 2, which is 8. So, there are 8 golf balls. Got that down. Now, out of these 8 golf balls, half are blue. Hmm, so again, half of 8 would be 8 divided by 2. Let me calculate that. 8 divided by 2 is 4. So, does that mean there are 4 blue golf balls? That seems right, but let me double-check to make sure I didn't skip any steps or misunderstand the question.

Wait, the problem says half of the balls are golf balls. So total balls: 16. Half are golf balls: 8. Then half of the golf balls are blue. So half of 8 is indeed 4. So, 4 blue golf balls. Let me just visualize this. If there are 16 balls in total, and half (so 8) are golf balls, and of those 8 golf balls, half are blue and half are another color. So splitting 8 into two equal parts gives 4 and 4. So yes, 4 blue golf balls. That makes sense.

Is there another way to approach this problem? Maybe by using fractions. Let's see. Total balls = 16. Half are golf balls: so 1/2 of 16. Then half of those are blue: so 1/2 of 1/2, which is 1/4. So 1/4 of the total balls are blue golf balls. 1/4 of 16 is 4. Yep, same answer. So whether I do it step by step or use fractions, I still get 4. That seems consistent.

Alternatively, if I were to write equations, let's denote total balls as T = 16. Number of golf balls G = T / 2. Then number of blue golf balls B = G / 2. Substituting, B = (16 / 2) / 2 = 8 / 2 = 4. Same result. So all methods lead to 4.

I think that's solid. Let me just check if I interpreted "half of the golf balls are blue" correctly. The problem didn't mention anything about the other half, but since half of the golf balls are blue, the other half could be any other color, but that's irrelevant to the question. The question only asks for the number of blue golf balls, so 4 is the answer.

Wait, just to be thorough, the problem doesn't have any trick wording, right? Like, when it says half of the balls are golf balls, that's straightforward. Then half of those golf balls are blue. So no, no tricky parts. If it had said half of the blue balls are golf balls, that would be different, but it's the other way around. Half of the golf balls are blue. So straightforward.

Another check: total golf balls are 8. If 4 are blue, then 4 are not blue. That adds up to 8. Then the other 8 balls (non-golf balls) are something else entirely. But since the question is only about blue golf balls, we don't need to consider those.

I think that's all. All angles checked. So, 4 blue golf balls. Confident that's correct.

\textbf{Final Answer:} The number of blue golf balls is \boxed{4}.
\end{tcolorbox}
\caption{A detailed example of mathematical problem-solving from R1 with a full reasoning trace.}
\label{fig:trace-example}
\end{figure*}


\section{Emergent Trace Analysis}
\label{sec:app_tracean}
This appendix provides quantitative details from the analysis performed on 16,710 successful emergent reasoning traces generated by R1 model, taken from~\citet{bespoke_stratos}. 
This analysis informs the discussion of pivot types and reasoning stages in Section~\ref{sec:structure_pivots}.
On average, these traces are substantial in length, comprising 2,713 tokens. 
Structurally, they contain an average of 141.7 paragraphs. 
Metacognitive activity is frequent, with an average of 93.5 identified pivots occurring within each reasoning trace.

\subsection{Details of Categorizing the Reasoning Trace Contents}
To analyze reasoning traces, we developed a systematic approach for identifying pivots and reasoning stages using regular expression pattern matching. 
For each of the four pivot categories (Realization, Verification, Exploration, and Integration), we created comprehensive regex patterns capturing lexical markers that signal metacognitive transitions -- for instance, phrases such as ``Actually'' for Realization pivots or ``Therefore'' for Integration pivots. 
Similarly, we defined patterns for the four reasoning stages (Problem Framing, Exploration, Verification, and Synthesis) based on characteristic expressions and structural elements. 
We experimented with using an LM in the form of GPT-4o mini for this task but found it to be no more performant.
This framework enabled automated extraction and quantification of reasoning elements across all 16,710 traces. 
While matching to regular expressions is not a perfect way to extract this information, we found it to perform well considering the very standardized nature of reasoning traces.
We iteratively refined our patterns over multiple rounds after manual inspection of reasoning traces by the authors, ensuring balanced detection across all categories.
Each trace was analyzed for both the frequency of pivot occurrences and the presence of reasoning stages, allowing us to quantify both the metacognitive transitions and the structural patterns that characterize effective reasoning.

\subsection{Pivot Analysis}
\noindent\textbf{Pivot Diversity}
A key indicator of complex reasoning is the variety of metacognitive shifts employed. 
The analyzed traces show an average diversity of 3.51 distinct pivot types per trace. 
Furthermore, a very high majority, 96.1\% of the traces, contains at least three different pivot categories. 
This high percentage strongly confirms the observation that successful traces typically involve multiple forms of reflection and correction during the reasoning process.

\noindent\textbf{Pivot Type Frequencies}
The frequency and prevalence of each specific pivot type across the dataset are summarized in Table~\ref{tab:pivot_frequencies}. 
Notably, Integration and Realization pivots remain extremely common, appearing in nearly all analyzed traces. 
Exploration pivots are the least prevalent, present in 87.0\% of traces, indicating consideration of alternative paths or hypotheses does not occur as frequently.

\begin{table}[h!]
\centering
\begin{tabular}{l r r}
\toprule
\textbf{Pivot Type} & \textbf{Avg. Occurrences per Trace} & \textbf{\% Traces Present} \\
\midrule
Realization & 18.96 & 98.6\% \\
Exploration & 16.11 & 87.0\% \\
Verification & 1.37 & 89.6\% \\
Integration & 67.64 & 100.0\% \\
\bottomrule
\end{tabular}
\caption{Frequency and prevalence of identified pivot types within the analyzed emergent reasoning traces (N=16,710).}
\label{tab:pivot_frequencies}
\end{table}

\subsection{Reasoning Stage Analysis}
The analysis also quantifies the presence of segments corresponding to the four cognitive reasoning stages discussed in ~\cref{sec:structure_pivots}. 
The difference compared to the pivot types analysis is that a single reasoning stage can contain multiple pivots, even of separate types.
In this case, we ascribe the trace to the reasoning stage found at the beginning of the trace.
The average occurrences and prevalence of these stages are detailed in ~\cref{tab:stage_frequencies}. 
Synthesis stages, often comprising multiple steps or integration points, are universally present. 
Problem Framing, Verification, and Exploration stages are all highly prevalent, appearing in the vast majority of traces, consistent with the high frequency of their corresponding pivot types.

\begin{table}[h!]
\centering
\begin{tabular}{l r r}
\toprule
\textbf{Reasoning Stage} & \textbf{Avg. Occurrences per Trace} & \textbf{\% Traces Present} \\
\midrule
Problem Framing & 3.13 & 79.2\% \\
Exploration & 6.81 & 87.0\% \\
Verification & 3.34 & 89.6\% \\
Synthesis & 84.08 & 100.0\% \\
\bottomrule
\end{tabular}
\caption{Frequency and prevalence of identified reasoning stages within the analyzed emergent reasoning traces (N=16,710).}
\label{tab:stage_frequencies}
\end{table}

\subsection{Key Insights from Analysis}
The quantitative analysis reinforces the qualitative observations presented in the main paper. 
Successful emergent reasoning traces consistently utilize a diverse range of pivot types, averaging 3.51 distinct types per trace, with an overwhelming 96.1\% using three or more, indicating frequent and varied metacognitive adjustments.
Furthermore, the presence and frequency of identified reasoning stage segments, detailed in Table~\ref{tab:stage_frequencies}, generally align with the cognitive science framework encompassing problem framing, exploration, verification, and synthesis.
The high prevalence of all four stage types underscores the iterative nature of the observed reasoning process.


\begin{table}[h!] 
\centering
\begin{tabular}{l r}
\toprule
\textbf{Data Source} & \textbf{Number of Samples} \\
\midrule
OlympicArena~\citep{huang2024olympicarenabenchmarkingmultidisciplinecognitive} & 4,250 \\
AGIEval~\citep{zhong2023agievalhumancentricbenchmarkevaluating} & 2,385 \\
LiveCodeBench v4~\citep{jain2024livecodebench} & 713 \\
NuminaMATH~\citep{numina_math_datasets} & 20,000 \\
OmniMath~\citep{gao2024omnimathuniversalolympiadlevel} & 4,238 \\
\midrule
\textbf{Total} & \textbf{31,586} \\
\bottomrule
\end{tabular}
\caption{Composition of the seed data pool used for generating reasoning traces. The number of samples from NuminaMATH reflects a randomly selected subset.}
\label{tab:seed-data}
\end{table}

\section{Dataset Statistics}
\label{sec:appendix} 
This section provides detailed statistics about the datasets created and utilized in our study, namely the seed data pool and the derived reasoning trace datasets \textsc{SmolTraces} (\textsc{ST}) and \textsc{SmolTraces-HardCoded} (\textsc{ST-HC}).

\subsection{Seed Data Compilation}
\label{subsec:app_seed_data} 
The foundation for generating our reasoning trace datasets is a curated collection of question-answer pairs sourced from diverse benchmarks spanning mathematics, science, logic, and coding. As detailed in Section~\ref{sec:datagen}, we selected problems from OlympicArena~\citep{huang2024olympicarenabenchmarkingmultidisciplinecognitive}, AGIEval~\citep{zhong2023agievalhumancentricbenchmarkevaluating}, LiveCodeBench v4~\citep{jain2024livecodebench}, NuminaMATH~\citep{numina_math_datasets}, and OmniMath~\citep{gao2024omnimathuniversalolympiadlevel}. This process resulted in a final seed dataset comprising 31,586 unique question-answer pairs. Table~\ref{tab:seed-data} presents a breakdown of the sources contributing to this seed data pool. To ensure the integrity of our downstream evaluations, we performed decontamination on this initial pool by removing any questions overlapping with our chosen evaluation benchmarks (MATH500, AIME2024, GPQA Diamond).

\begin{table}[h]
\centering
\begin{tabularx}{\linewidth}{@{} l X @{}} 
\toprule
\textbf{Statistic} & \textbf{Dataset Values} \\ 
\midrule
Trace Origin & 
    \textsc{ST}: Emergent (R1) \newline 
    \textsc{ST-HC}: Synthetic (GPT-4o + Hard-coded Prompt, Fig.~\ref{fig:prompt}) \\
\midrule
Seed Questions Source & 
    \textsc{ST}: Shared Pool (Table~\ref{tab:seed-data}, N=31,586) \newline 
    \textsc{ST-HC}: Shared Pool (Table~\ref{tab:seed-data}, N=31,586) \\
\midrule
Final Number of Samples & 
    \textsc{ST}: 25,802 (adjusted to N=18,242 after) \newline
    \textsc{ST-HC}: 18,242 \\ 
\midrule
Filtering Applied & 
    \textsc{ST}: Correct Answer (up to 5 attempts), Min. Length (50 tokens) \newline 
    \textsc{ST-HC}: Correct Answer (up to 5 attempts), Min. Length (50 tokens) \\
\midrule
Avg. Trace Length (Tokens) & 
    \textsc{ST}: 2,521 \newline 
    \textsc{ST-HC}: 2,101 \\
\midrule
Avg. Pivots per Trace & 
    \textsc{ST}: 93.4 \newline 
    \textsc{ST-HC}: 89.1 \\
\bottomrule
\end{tabularx}
\caption{Summary statistics for the final generated reasoning trace datasets used in fine-tuning. Values reflect the approximately 18,250 filtered and balanced samples in each dataset (\textsc{ST} and \textsc{ST-HC}). Average length and pivot counts are illustrative; actual values depend on the final composition.}
\label{tab:dataset_summary_appendix}
\end{table}

\subsection{Generated Reasoning Trace Datasets}
\label{subsec:app_generated_datasets}

Using the curated seed data, we generated two parallel datasets featuring detailed reasoning traces, as described in Section~\ref{sec:datagen}. Table~\ref{tab:dataset_summary_appendix} provides a summary comparing key statistics of the final, balanced \textsc{ST} and \textsc{ST-HC} datasets. Average trace length and pivot counts reflect measurements across these final samples.

\noindent\textbf{\textsc{SmolTraces (ST)}:} This dataset contains emergent reasoning traces generated by the state-of-the-art RLM R1~\citep{deepseekai2025deepseekr1incentivizingreasoningcapability}. For each seed question, we prompted R1 in a zero-shot manner up to five times, retaining the first trace that yielded the correct final answer.

\noindent\textbf{\textsc{SmolTraces-HardCoded (ST-HC)}:} This dataset comprises synthetic reasoning traces generated using GPT-4o, guided by the structured prompt detailed in \cref{fig:prompt} (Figure~\ref{fig:prompt}). This prompt enforces the inclusion of specific structural elements and lexical pivots identified in our analysis (Section~\ref{sec:structure_pivots}). Similar to \textsc{ST}, generation involved up to five zero-shot attempts per seed question, keeping the first correct trace.

\noindent\textbf{Balancing:} To ensure a fair comparison in finetuning experiments, we balanced the datasets by size. 
As the \textsc{ST-HC} dataset contained fewer samples after filtering, we downsample the larger \textsc{ST} dataset by randomly removing samples that are not present in the smaller dataset until both datasets have an equal number of samples.
Therefore, the final versions of both \textsc{ST} and \textsc{ST-HC} used for finetuning contain an equal number of samples (N=18,242).


\section{Synthetic Traces with Wrong Answer Details}
\label{sec:app_abla}
This section details the construction process for the \textsc{ST-HC-W} dataset, used in our ablation study (Section~\ref{sec:abl}) to investigate the impact of reasoning trace style independent of final answer correctness. The goal is to create a dataset that retains the stylistic and structural characteristics of the synthetic \textsc{ST-HC} traces but deliberately leads to an incorrect final answer. First, we use the ground truth answer associated with the seed question for each sample in \textsc{ST-HC}. Then, we prompt GPT-4o-mini, instructing it to provide a different, incorrect answer that is similar in format (e.g. ``9.11'' instead of ``9.9''). We compare the answer generated by GPT-4o-mini against the known correct answer and if the generated answer matches the correct one, we repeat the procedure.

The resulting \textsc{ST-HC-W} dataset mirrors the size of \textsc{ST-HC}, containing 18,242 samples. Each sample includes the original question, a reasoning trace stylistically similar to \textsc{ST-HC}, but which concludes with an incorrect final answer. This construction allows for better isolation of the effect of learning stylistic reasoning patterns during finetuning.

\end{document}